\documentclass{article}
\usepackage{spconf,amsmath,graphicx}

\usepackage{enumitem}

\usepackage{hyperref}
\usepackage{epsfig}
\usepackage{graphicx}
\usepackage{url}
\usepackage{color}
\usepackage{mathrsfs}  
\usepackage{amssymb,amsmath,bm}
\usepackage{multicap,graphicx}
\usepackage{caption}
\usepackage{algorithm}  
\usepackage{algorithmic}
\usepackage{graphicx}
\usepackage{wrapfig}
\usepackage{picinpar}
\usepackage{cutwin}
\usepackage{stmaryrd}

\definecolor{amethyst}{rgb}{0.54, 0.17, 0.89}
\title{Enhanced Separable Disentanglement for \\ Unsupervised Domain Adaptation}
\name{Youshan Zhang \ \ \ \ \ \ \ \  Brian D.\ Davison}
\address{Computer Science and Engineering, Lehigh University, Bethlehem, PA, USA \\ \{yoz217, bdd3\}@lehigh.edu}

\begin{document}
%
\maketitle
\begin{abstract}
Domain adaptation aims to mitigate the domain gap when transferring knowledge from an existing labeled domain to a new domain.  However, existing disentanglement-based methods do not fully consider separation between domain-invariant and domain-specific features, which means the domain-invariant features are not discriminative. The reconstructed features are also not sufficiently used during training. In this paper, we propose a novel enhanced separable disentanglement (ESD) model. We first employ a disentangler to distill domain-invariant and domain-specific features. Then, we apply feature separation enhancement processes to minimize contamination between domain-invariant and domain-specific features. Finally, our model reconstructs complete feature vectors, which are used for further disentanglement during the training phase. Extensive experiments from three benchmark datasets outperform state-of-the-art methods, especially on challenging cross-domain tasks.
\end{abstract}
\section{Introduction}
\label{sec:intro}
Most existing machine learning models rely on large amounts of labeled training data to achieve high performance.  Unfortunately, such a requirement cannot be met in many real-world applications.   The number of labels is limited and manual annotation is expensive and time-consuming. Therefore, it is valuable to learn a model for a new domain from one with existing labeled samples. However, the difference between the two domains, termed the domain shift, can cause difficulty with direct use of models from one domain on another. Unsupervised domain adaptation (UDA) has emerged as a prominent method to address the domain shift.


Deep learning models have been widely used in UDA. Earlier methods rely on minimizing the discrepancy between the source and target distributions by proposing different loss functions, such as Maximum Mean Discrepancy (MMD)~\cite{tzeng2014deep}, CORrelation ALignment~\cite{sun2016deep}, Kullback-Leibler divergence~\cite{meng2018adversarial}. To learn domain invariant features, adversarial domain adaptation methods aim to identify domain invariant features by playing a min-max game between domain discriminator and feature extractor~\cite{ghifary2014domain,tzeng2017adversarial}. However, most of these UDA methods do not consider constructing separable domain-specific features ($f_{ds}$) and domain-invariant features ($f_{di}$) to learn more discriminative representations.

Recently, disentanglement representation based methods can learn discriminative $f_{ds}$, which contains domain related information, and $f_{di}$, which contains intrinsic information related to different categories. Cross domain representation disentangler~\cite{liu2018detach} bridged labeled source domain and unlabeled target domain via jointly learning cross-domain feature representation disentanglement and adaptation.  Gonzalez et al.~\cite{gonzalez2018image} proposed an image-to-image translation for representation disentangling based on GANs and cross-domain autoencoders. They separated the internal representation into three parts: shared part, which contains the domain invariant features for two domains, and two exclusive parts, which contain the domain-specific features.  Peng et al.~\cite{peng2019domain} minimized mutual information between $f_{ds}$ and $f_{di}$ to pursue implicit domain invariant features.  Liu et al.~\cite{liu2018unified} introduced a unified feature disentanglement network to learn a domain-invariant representation from multiple domains for image translation and manipulation. Gholami et al.~\cite{gholami2020unsupervised} presented a multi-target domain adaptation information theoretic approach to find a shared latent space of all domains by simultaneously identifying the remaining private, domain-specific factors.  However, these methods either cannot fully separate $f_{ds}$ and $f_{di}$~\cite{liu2018detach,gonzalez2018image,liu2018unified} or the reconstructed features are 
insufficiently used during training to facilitate the performance of a domain discriminator and a domain-invariant classifier~\cite{peng2019domain,gholami2020unsupervised}.  

\begin{figure*}[t]
\centering
\includegraphics[scale=0.3]{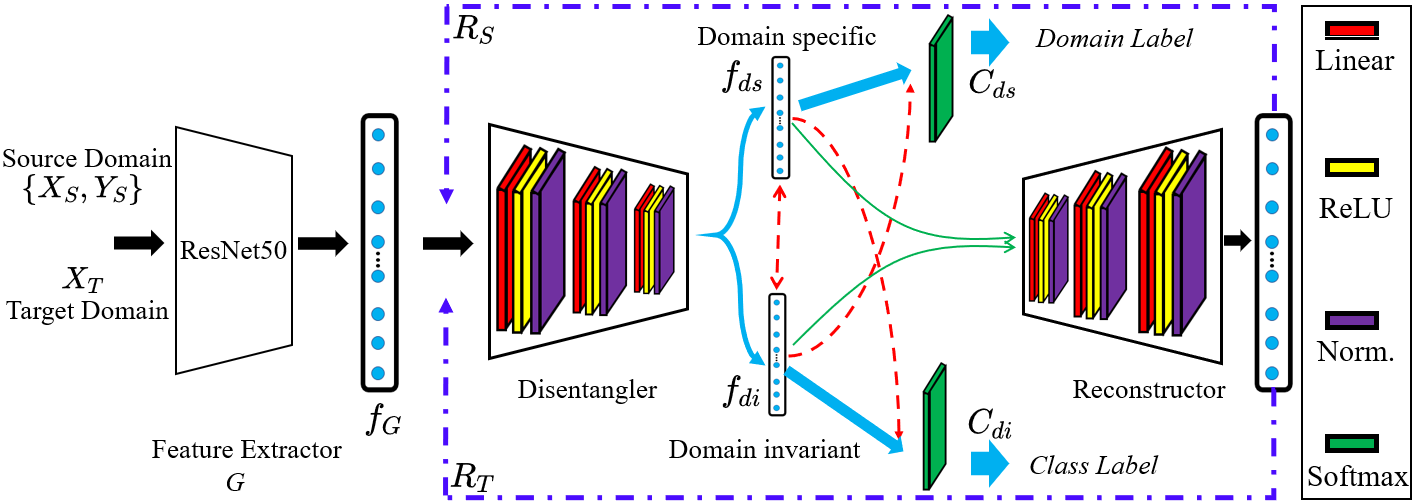}
\caption{An overview of ESD. We first employ  ResNet50 as feature extractor $G$ to extract features of two domains. We then employ 1) a disentangler to distill domain-specific $f_{ds}$ and domain-invariant $f_{di}$ features and then train domain-specific classifier $C_{ds}$ and invariant classifier $C_{di}$, respectively (\textcolor{cyan}{blue arrows});  2) feature separation processes to enhance the disentangler and minimize contamination between $f_{ds}$ and $f_{di}$ (\textcolor{red}{red lines}); 3) a reconstructor to recover original feature prototypes (\textcolor{green}{green lines}) and finally re-utilize the reconstructed features ($R_S$ and $R_T$) to improve $C_{ds}$ and $C_{di}$ (\textcolor{amethyst}{purple lines}). Norm.: normalization.}
\label{fig:ESD}
\vspace{-0.3cm}
\end{figure*}

To alleviate these issues, we propose an enhanced separable disentanglement (ESD)  model. Our contributions are:

\begin{itemize}[noitemsep,topsep=0pt]
    \item We propose a novel method for feature disentanglement representation learning: 1) teach a disentangler to distill domain-specific from domain-invariant features; 2) apply feature separation maximization processes to enhance the disentangler and improve the effectiveness of both kinds of features; 3) design a reconstructor to recover original feature prototypes which can be further re-utilized in steps 1) and 2) to improve performance.   
  
    \item For feature separation maximization, we first propose a novel structural similarity loss to maximize the dissimilarity between $f_{ds}$ and $f_{di}$, and then propose opposite binary cross-entropy loss and accurate loss to further remove contaminated information.
    \item  The reconstructor recovers original feature prototypes using reconstructed features which are then used to update the learned classifier.
\end{itemize} 
Extensive experiments on three benchmark datasets show that our ESD model outperforms state-of-the-art methods.

\section{Methods}
\vspace{-.1in}
{\bf Problem and notation.}
For UDA, given a source domain $\mathcal{D_S} = \{X_{S}^i, Y_{S}^i \}_{i=1}^{n_s}$ of $n_s$ labeled samples in $K$ categories and a target domain $\mathcal{D_T} = \{X_{T}^j\}_{j=1}^{n_t}$ without any labels ($Y_T$ for evaluation only), our ultimate goal is to learn a classifier under a feature extractor $G$, that produces lower generalization error in the target domain.

{\bf Disentangler.}
As shown in Fig.~\ref{fig:ESD}, feature extractor $G$ maps a labeled source domain and an unlabeled target domain into latent feature prototypes $f_G$. For a classical disentangler (D), it disentangles $f_G$ into domain-specific features $f_{ds}$, and domain-invariant features $f_{di}$. In the first step, for $f_{di}$, we aim to train a domain invariant classifier $C_{di}$ using the labeled source invariant features $f_{di}^S$ and make predictions for the target invariant features $f_{di}^T$ with typical cross-entropy loss as follows:
\begin{equation}\label{eq:lc}
    \mathcal{L}_{di} = \mathcal{L_C}(C_{di}(f_{di}^S), Y_S), 
\end{equation}
where $\mathcal{L_C}$ is cross-entropy loss. 
For $f_{ds}$, we aim to learn a domain discriminator classifier $C_{ds}$ using adversarial loss to distinguish the source domain-specific features $f_{ds}^S$ and the target domain-specific features $f_{ds}^T$ in the following equation.
\begin{equation}\label{eq:l_ds}
    \mathcal{L}_{ds}  = \mathbb{E}_{x_s^i \sim f_{ds}^S }  \text{log}[ C_{ds} (x_s^i )] +  \mathbb{E}_{x_t^j \sim f_{ds}^T }  \text{log}[ 1-C_{ds} (x_t^j )],
\end{equation}
where domain labels of $f_{ds}^S$ and $f_{ds}^T$ are set as 1 and 0. However, a traditional disentangler cannot ensure that there is no contaminated information between $f_{di}$ and $f_{ds}$. We employ a feature separation enhancement step to alleviate this issue.

{\bf Feature Separation Enhancement.}
In step two, to guarantee the completed feature separation between $f_{di}$ and $f_{ds}$, we also employ three processes. First of all, we maximize the dissimilarity between $f_{di}$ and $f_{ds}$. Peng et al.~\cite{peng2019domain} proposes to minimize the mutual information between them. However, there is no closed solution for minimizing mutual information, and the Monte Carlo sampling will lead to external computation. Instead, we directly maximize the dissimilarity between the disentangled features, which is equivalent to minimizing the similarity between $f_{di}$ and $f_{ds}$. We impose a batch-wise structural similarity loss. 
\begin{equation}
\begin{aligned}
\mathcal{L}_{S}= abs(\frac{(2\mu_{B_1} \mu_{B_2} +C_1)(2\sigma_{{B_1}{B_2}} +C_2)}{(\mu_{B_1}^2+\mu_{B_2}^2+C_1)(\sigma_{B_1}^2+\sigma_{B_2}^2+C_2)}), 
\end{aligned}
\end{equation}
where $abs$ takes the absolute value, $B_1 \in f_{di}$ and $B_2 \in f_{ds}$ are batch-wise features, $\mu_{B_1},  \mu_{B_2}, \sigma_{B_1} ,\sigma_{B_2}$, and $\sigma_{{B_1}{B_2}}$ are mean, standard deviations of domain invariant and specific features batch, and cross-covariance for $(B_1, B_2)$. $C_1$ and $C_2$ are two variables to stabilize the division with weak denominator. This loss function is derived from structural similarity index measure (SSIM)~\cite{wang2004image}. It has the advantages of measuring luminance, contrast and structural difference between $B_1$ and $B_2$. Therefore, $\mathcal{L}_{S}$ has more capability of measuring the similarity between $f_{di}$ and $f_{ds}$. In addition, the range of the $\mathcal{L}_{S}$ is from $0$ to $1$, where $1$ indicates high similarity between batch features, and $0$ means they are not similar. During the training, we keep minimizing such a similarity, and thus contamination between $f_{di}$ and $f_{ds}$ is also minimized.

Secondly, to further ensure domain-specific features are fully segregated from domain-invariant features, we leverage $f_{di}$ to fool the trained domain-specific classifier $C_{ds}$ using opposite binary cross-entropy loss in Eq.~\ref{eq:l_o}.  
\begin{equation}\label{eq:l_o}
    \mathcal{L}_{O}  = \mathbb{E}_{x_s^i \sim f_{di}^S }  \text{log}[1- C_{ds} (x_s^i )] +  \mathbb{E}_{x_t^j \sim f_{di}^T }  \text{log}[C_{ds} (x_t^j )]
\end{equation}
Differing from Eq.~\ref{eq:l_ds}, the labels of $f_{di}^S$ and $f_{di}^T$ are set as 0 and 1, oppositely. Since we want the trained $C_{ds}$ to be unable to predict the correct domain labels of $f_{di}$, we effectively force $f_{ds}$ to contain no $f_{di}$ information.

Thirdly, to remove $f_{ds}$ from $f_{di}$, we design an accurate loss to force no correct labels can be predicted using trained domain invariant classifier $C_{di}$ as follows:
\begin{equation}\label{eq:l_a}
    \mathcal{L}_{A}  = \frac{|C_{di}(f_{ds}^S) \wedge C_{di}(f_{di}^S)|}{n_s} + \frac{|C_{di}(f_{ds}^T) \wedge C_{di}(f_{di}^T)|}{n_t},
\end{equation}
where $\wedge$ is the and operation, and $|\cdot|$ measures the length. The numerator measures the number of the same predictions of source and target $f_{ds}$ and $f_{di}$, respectively. Minimizing such an accurate loss, we can make sure that the predictions of $f_{ds}$ are different from $f_{di}$.  Hence, we can force $f_{di}$ contains no $f_{ds}$ information.

Considering the above three processes, we can enhance the disentangler to minimize the contamination between $f_{di}$ and $f_{ds}$.

{\bf Reconstructor.}
In step three, to keep the information integrity of feature disentanglement, we train a reconstructor ($R$) to recover original feature prototypes ($f_G$)  using the disentangled $f_{di}$ and $f_{ds}$. Let $R_S = R(f_{di}^S, f_{ds}^S)$ and $R_T = R(f_{di}^T, f_{ds}^T)$, the reconstruction loss is defined as:
\begin{equation}
\begin{aligned}\label{eq:R}
    \mathcal{L}_{R} =  ||R_S - f_G^S||_2^2 + ||R_T - f_G^T||_2^2.
\end{aligned}
\end{equation}
By this reconstructor R, we can guarantee that $R_S \sim f_G^S$ and $R_T \sim f_G^T$. Differing from previous work, we take advantage of reconstructed features to further improve the performance of $C_{di}$ and $C_{ds}$. As shown in purple lines in Fig.~\ref{fig:ESD}, we re-utilize the reconstructed features in the disentangler. At this stage, all parameters in step one and two are fixed, \textit{i.e.}, $D, C_{di}$ and $C_{ds}$ are fixed. We directly apply the trained disentangler D to get the disentangled features $\hat{f_{di}}$ and $\hat{f_{ds}}$ using the reconstructed features $R(f_G)$. Therefore, we will repeat the minimization equations in step one and step two in Eq.~\ref{eq:12}. 
\begin{equation}
\begin{aligned}\label{eq:12}
    \mathcal{L}_{T} =  \hat{\mathcal{L}_{di}} + \hat{\mathcal{L}_{ds}} + \hat{\mathcal{L}_{S}} + \hat{\mathcal{L}_{O}} + \hat{\mathcal{L}_{A}},
\end{aligned}
\end{equation}
where in Eq.~\ref{eq:lc} to Eq.~\ref{eq:l_a}, we will directly compute the loss by replacing $f_{di}$ and $f_{ds}$ with $\hat{f_{di}}$ and $\hat{f_{ds}}$, and replacing $f_{di}^S, f_{di}^T, f_{ds}^S, f_{ds}^T$ with $\hat{f_{di}^S}, \hat{f_{di}^T}, \hat{f_{ds}^S}, \hat{f_{ds}^T}$, while all other components are the same as before.

{\bf The overall training objective function.}
The architecture of our proposed ESD  model is shown in Fig.~\ref{fig:ESD}.  Our model minimizes the following objective function.
\begin{equation}
\begin{aligned}\label{eq:all}
\mathcal{L} & (X_S, Y_S, X_T)=  \mathop{\arg\min} \  ((\mathcal{L}_{di} + \mathcal{L}_{ds}) \\ & + \alpha (\mathcal{L}_{S} + \mathcal{L}_{O} + \mathcal{L}_{A}) + \beta (\mathcal{L}_{R} + \mathcal{L}_{T}))
\end{aligned}
\end{equation}
where $(\cdot)$ represents loss functions in each step, $\alpha$ and $\beta$ are factors to balance the importance of steps two and  three.

\section{Experiments}
\vspace{-.1in}
{\bf Datasets.}
We test our model using three public image datasets:  Office-31, Office-Home and VisDA-2017. \textbf{Office-31} \cite{saenko2010adapting} has 4,110 images from three  domains: Amazon (A), Webcam (W), and DSLR (D) in 31 classes. In experiments, A$\shortrightarrow$W represents transferring knowledge from domain A to domain W.  \textbf{Office-Home}~\cite{venkateswara2017deep} dataset contains 15,588 images from four domains: Art (Ar), Clipart (Cl), Product (Pr) and Real-World (Rw) in 65 classes. \textbf{VisDA-2017}~\cite{peng2017visda} is a challenging dataset due to the big domain-shift between the synthetic images (152,397 images from VisDA) and the real images (55,388 images from COCO) in 12 classes. We evaluate our method on the setting of synthetic-to-real as the source-to-target domain and report accuracy of each category. 

{\bf Implementation details.}
We implement our approach using PyTorch and extract features for the three datasets from  finely tuned ResNet50 (Office-31, Office-Home) and ResNet101 (VisDA-2017) networks~\cite{zhang2020adversarial}. The 1,000 features are then extracted from the last fully connected layer for the source and target features. In the Disentangler $D$, the number of outputs of the first two Linear layers are 1000 and 512, and the output of the last Linear layer is the number of classes ($K$) in each dataset, while the output in the reconstruction layers is opposite ($K$, 512, and 1000). $C_1 = 0.01^2$ and $C_2=0.03^2$ as in~\cite{wang2004image}. The learning rate = 0.001, batch size = 32 and number of iterations = 100 with SGD optimizer. The balanced factors $\alpha = 0.3$ and $\beta = 0.1$.

\begin{table*}[h!]
\begin{center}
\small
\caption{Accuracy (\%) on Office-Home dataset (based on ResNet50)}
\vspace{-0.3cm}
\setlength{\tabcolsep}{+1.1mm}{
\begin{tabular}{rccccccccccccc}
\hline \label{tab:OH}
Task & Ar$\shortrightarrow$Cl &  Ar$\shortrightarrow$Pr & Ar$\shortrightarrow$Rw & Cl$\shortrightarrow$Ar & Cl$\shortrightarrow$Pr & Cl$\shortrightarrow$Rw & Pr$\shortrightarrow$Ar & Pr$\shortrightarrow$Cl & Pr$\shortrightarrow$Rw & Rw$\shortrightarrow$Ar & Rw$\shortrightarrow$Cl & Rw$\shortrightarrow$Pr & \textbf{Ave.}\\
\hline
ResNet-50~\cite{he2016deep}&	34.9&	50.0&	58.0&	37.4&	41.9&	46.2&	38.5&	31.2&	60.4&	53.9&	41.2&	59.9&	46.1\\
DAN~\cite{long2015learning}	& 43.6	& 57.0& 	67.9& 	45.8& 	56.5& 	60.4& 	44.0& 	43.6& 	67.7& 	63.1& 	51.5& 	74.3& 	56.3\\
DANN~\cite{ghifary2014domain} 	& 45.6	& 59.3& 	70.1& 	47.0& 	58.5& 	60.9& 	46.1& 	43.7& 	68.5& 	63.2& 	51.8& 	76.8& 	57.6\\
JAN~\cite{long2017deep}		& 45.9& 	61.2& 	68.9& 	50.4& 	59.7& 	61.0& 	45.8& 	43.4& 	70.3& 	63.9& 	52.4& 	76.8& 	58.3\\
TAT~\cite{liu2019transferable} &51.6  &69.5 & 75.4 & 59.4 & 69.5 & 68.6 & 59.5 &50.5 &76.8 &70.9 &56.6 &81.6 &65.8 \\
ETD~\cite{li2020enhanced} &51.3 & 71.9& 85.7& 57.6 &69.2 &73.7 &57.8 &51.2 &79.3 &70.2 &57.5 &82.1 &67.3 \\
SymNets~\cite{zhang2019domain} & 47.7 & 72.9 & 78.5 & 64.2  & 71.3  &74.2  & 64.2  & 48.8 &  79.5&  74.5 &52.6 & 82.7& 67.6 \\
DCAN~\cite{li2020domain} &\textbf{54.5 }&75.7 &81.2 &67.4 &74.0 &76.3 &67.4 &52.7 &80.6 &74.1 &\textbf{59.1} &83.5 &70.5\\ 
\hline
\hline
\textbf{ESD}& 53.2 &	\textbf{75.9}  &	\textbf{82.0}  &	\textbf{68.4} &	\textbf{79.3}  & \textbf{79.4} & \textbf{69.2} &	\textbf{54.8} &	\textbf{81.9} &	\textbf{74.6} &	56.2  &	\textbf{83.8} & 	\textbf{71.6} \\
\hline
\end{tabular}}
\vspace{-0.6cm}
\end{center}
\end{table*}

\begin{table*}[h!]
\begin{center}
\small
\caption{Accuracy (\%) on VisDA-2017 dataset (based on ResNet101)}
\vspace{-0.3cm}
\setlength{\tabcolsep}{+2.1mm}{
\begin{tabular}{rccccccccccccc}
\hline \label{tab:VisDA}
Task & plane& bcycl& bus& car& horse& knife& mcycl& person& plant& sktbrd& train& truck & \textbf{Ave.}\\
\hline
Source-only~\cite{he2016deep} &  55.1 &53.3 &61.9& 59.1& 80.6& 17.9& 79.7& 31.2& 81.0& 26.5& 73.5& 8.5 & 52.4 \\
DANN~\cite{ghifary2014domain} 	&81.9 &77.7& 82.8 &44.3& 81.2& 29.5 &65.1 &28.6 & 51.9 & 54.6 & 82.8 & 7.8 & 57.4\\
DAN~\cite{long2015learning}	& 87.1& 63.0& 76.5& 42.0 &90.3& 42.9 &85.9 &53.1& 49.7 &36.3& 85.8 &20.7 &61.1\\
JAN~\cite{long2017deep}	& 75.7& 18.7& 82.3 &86.3& 70.2 &56.9& 80.5& 53.8 &92.5 &32.2& 84.5& 54.5 &65.7 \\
MCD~\cite{saito2018maximum}	& 87.0 &60.9& 83.7& 64.0& 88.9& 79.6& 84.7& 76.9& 88.6& 40.3& 83.0& 25.8& 71.9\\
DADA~\cite{tang2020discriminative} & 92.9 &74.2& 82.5& 65.0& 90.9& 93.8& 87.2& 74.2& 89.9& 71.5& 86.5 &48.7&  79.8 \\
STAR~\cite{lu2020stochastic} & 95.0& 84.0& 84.6& 73.0& 91.6 &91.8& 85.9 &78.4& 94.4& 84.7 &87.0 &42.2& 82.7 \\
CAN~\cite{kang2019contrastive} & \textbf{97.9} & 87.2& 82.5& 74.3 &\textbf{97.8} & 96.2& 90.8& 80.7& 96.6 &96.3 &87.5& 59.9& 87.2 \\
\hline
\hline
\textbf{ESD}&   96.8 & \textbf{89.1} & \textbf{87.9} & \textbf{80.3} & 96.7 & \textbf{96.9} & \textbf{92.5} & \textbf{84.9} & \textbf{96.9} & \textbf{97.5} & \textbf{88.9} & \textbf{62.8} & \textbf{89.2} \\
\hline
\end{tabular}}
\vspace{-0.9cm}
\end{center}
\end{table*}
\begin{table}[!htb]
\small
      \caption{Accuracy (\%) on Office-31 (based on ResNet50)}
      \vspace{-0.3cm}
      \centering
\setlength{\tabcolsep}{+0.6mm}{
\begin{tabular}{rcccccccc|c|c|c|c|c|c|c|c|}
\hline \label{tab:O31}
Task & A$\shortrightarrow$W &  A$\shortrightarrow$D & W$\shortrightarrow$A & W$\shortrightarrow$D & D$\shortrightarrow$A & D$\shortrightarrow$W  & \textbf{Ave.}\\
\hline
RTN~\cite{long2016unsupervised} &	84.5 &	77.5 &	64.8 &	99.4 &	66.2 &	96.8 &	81.6 \\
ADDA~\cite{tzeng2017adversarial}	&86.2	&77.8  &68.9	&98.4 &	69.5 &	96.2	& 82.9\\
JAN~\cite{long2017deep}&	85.4 &	84.7	&70.0 &	99.8	&68.6 &	97.4 &	84.3\\
TAT~\cite{liu2019transferable} & 92.5 & 93.2 & 73.1 & \textbf{100}& 73.1 & \textbf{99.3} & 88.4 \\
TADA~\cite{wang2019transferable} &94.3 & 91.6  & 73.0  &99.8  & 72.9 & 98.7 & 88.4 \\
SymNets~\cite{zhang2019domain}& 90.8& 93.9  &72.5  & \textbf{100} & 74.6 & 98.8& 88.4 \\
CAN~\cite{kang2019contrastive} & 94.5 & 95.0  &77.0   &99.8  & 78.0 & 99.1 &  90.6 \\
\hline
\hline
\textbf{ESD} & \textbf{94.6} &	\textbf{96.2} &	\textbf{80.1} &	99.0 &	\textbf{80.4} &	98.0 &	\textbf{91.4}	\\
\hline
\end{tabular}}
\vspace{-0.6cm}
\end{table}

\begin{table}[b]  
\begin{center}
\vspace{-0.3cm}
\caption{Ablation tests on Office-31  (minus steps II and III).}
\vspace{-0.2cm}
\setlength{\tabcolsep}{+0.3mm}{
\begin{tabular}{llllllll}
\hline \label{tab:ab}
Task & A$\shortrightarrow$W &  A$\shortrightarrow$D & W$\shortrightarrow$A & W$\shortrightarrow$D & D$\shortrightarrow$A & D$\shortrightarrow$W  & \textbf{Ave.}\\
\hline
ESD$-$II$-$III  & 91.2 &	87.6 &	76.3 &	98.2 &	76.2 &	96.6 &	87.7\\
ESD$-$II & 92.1 &	93.9 &	78.5 &	98.6&	79.1 &	96.6 &	89.8\\
ESD$-$III & 94.3 &	95.3 &	79.8 &	98.8 &	79.3 &	97.2 &	90.8	\\
\hline
\hline
\textbf{ESD} & \textbf{94.6} &	\textbf{96.2} &	\textbf{80.1} &	\textbf{99.0} &	\textbf{80.4} &	\textbf{98.0} &	\textbf{91.4}	\\
\hline
\end{tabular}}
\end{center}
\vspace{-0.5cm}
\end{table}

{\bf Comparison to state-of-the-art methods.}
We compare the performance of our ESD model with 15 state-of-the-art methods. For a fair comparison, baseline results are directly reported from their original papers. From Tables~\ref{tab:OH} to \ref{tab:O31}, we can observe that the accuracy of the ESD model is ahead of all other methods in most tasks. For the Office-31 dataset, the average accuracy of ESD is 91.4\%. It is superior to all other methods. ESD shows substantially better transferring ability than other methods in tasks  W$\shortrightarrow$A and D$\shortrightarrow$A. In another difficult Office-Home dataset, which has more categories, more samples, and larger domain discrepancy than the Office-31 datasets, the average accuracy is 71.6\%, which exceeds the SOTA methods. Our model has a particularly obvious improvement in the challenging VisDA-2017 dataset, which has many more samples and larger domain discrepancy than the other two datasets. Our model achieves the highest accuracy 89.2\%, which is two percent higher than the best baseline CAN. By testing on three distinct datasets, we give evidence of ESD's broad applicability.

\section{Discussion}
\vspace{-.1in}
 
In all experiments, our method achieves the highest average accuracy. 
There are two prominent reasons for the success of our model. First of all, the proposed three novel loss functions ($\mathcal{L}_S$, $\mathcal{L}_O$ and $\mathcal{L}_A$) in the feature separation enhancement step are important to remove contaminated information between domain-specific and domain-invariant features, which improves the performance of $C_{di}$.  Secondly, re-utilizing reconstructed features is also helpful in optimizing $C_{di}$ and $C_{ds}$ and leads to higher accuracy. We also observe that our model is suboptimal in some tasks (Rw$\shortrightarrow$Cl in Office-Home and D$\shortrightarrow$W in Office-31 dataset),
so we cannot guarantee that our model always beats all other methods. 

\begin{figure}[t]
\centering
\includegraphics[scale=0.3]{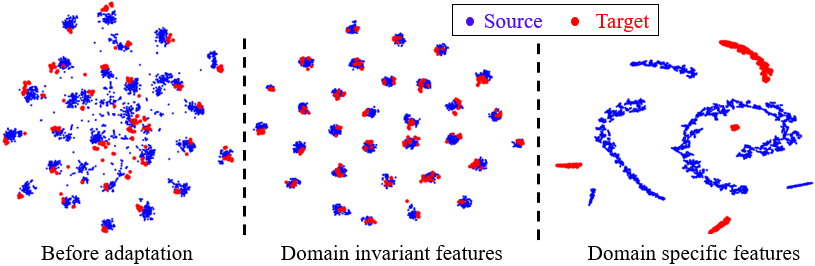}
\caption{Visualization of learned features using a 2D t-SNE view of task A$\shortrightarrow$D in Office-31 dataset.}
\label{fig:t-sne}
\vspace{-0.4cm}
\end{figure}

To show how different steps affect final performance, we also conduct an ablation study in Tab.~\ref{tab:ab} (step one is required). We find step two is more important than step three in improving final accuracy (90.8\% v.s.\ 89.8\%), but both contribute to final performance. To intuitively present adaptation performance and the effects of step two, we utilize t-SNE~\cite{van2008visualizing} to visualize the deep features of network activations in 2D space before and after adaptation (domain invariant/specific features) as shown in Fig.~\ref{fig:t-sne}. Apparently, the distributions of domains A and D become more discriminative after adaptation ($f_{di}$), while many categories are mixed in the feature space before adaptation. In addition, the $f_{ds}$ can also distinguish the target domain from the source domain. Furthermore, the distributions of $f_{di}$ and $f_{ds}$ are different, which implies that  contamination between them is minimized. This result indicates that ESD can learn more discriminative representations.

\vspace{-0.2cm}
\section{Conclusion}
\vspace{-0.1cm}
We have presented a novel enhanced separable disentanglement model to improve the separability between domain-specific  and domain-invariant features. Specifically, we impose structural similarity loss, opposite binary cross-entropy loss, and accurate loss functions to minimize contamination among disentangled features. We further re-utilize the reconstructed features to improve the performance of domain discrimination and domain-invariant classification. Extensive experiments demonstrate that our ESD model outperforms state-of-the-art methods.

\small
\bibliographystyle{IEEEbib}
\bibliography{strings,refs}

\end{document}